\documentclass[10pt, a4paper]{article}
\usepackage{lrec}
\usepackage{multirow}
\usepackage{times}
\usepackage{latexsym}
\usepackage{url}
\usepackage{comment}
\usepackage{arabtex}
\usepackage{utf8}
\usepackage{tikzsymbols}
\usepackage{amsmath}
\usepackage{graphicx}
\usepackage{multirow}
\usepackage{array,booktabs,makecell}
\usepackage{url}
\usepackage{enumitem}
\setenumerate[0]{label=\arabic*)}
\newcounter{examples}
\usepackage{wasysym}

\usepackage{multibib}
\newcites{languageresource}{Language Resources}
\usepackage{graphicx}
\usepackage{tabularx}
\usepackage{soul}

\usepackage{epstopdf}
\usepackage[utf8]{inputenc}

\usepackage{hyperref}
\usepackage{xstring}

\usepackage{color}

\title{Leveraging Affective Bidirectional Transformers for Offensive Language Detection}

\author{Muhammad Abdul-Mageed \\
   Natural Language Processing Lab / Address line 1 \\
   Affiliation / Address line 2 \\
   Affiliation / Address line 3 \\
   \texttt{email@domain} \\\And
   Second Author \\
   Affiliation / Address line 1 \\
   Affiliation / Address line 2 \\   Affiliation / Address line 3 \\
   \texttt{email@domain} \\}

\name{AbdelRahim Elmadany, Chiyu Zhang, Muhammad Abdul-Mageed, Azadeh Hashemi \\
 \texttt{\{a.elmadany,muhammad.mageeed,azadeh.hashemi\}@ubc.ca, chiyuzh@mail.ubc.ca}}

\address{Natural Language Processing Lab\\ University of British Columbia}

\date{}

\abstract{
Social media are pervasive in our life, making it necessary to ensure safe online experiences by detecting and removing offensive and hate speech. In this work, we report our submission to the Offensive Language and hate-speech Detection shared task organized with the 4\textsuperscript{th} Workshop on Open-Source Arabic Corpora and Processing Tools Arabic (OSACT4). We focus on developing purely deep learning systems, without a need for feature engineering. For that purpose, we develop an effective method for automatic data augmentation and show the utility of training both offensive and hate speech models off (i.e., by fine-tuning) previously trained affective models (i.e., sentiment and emotion). Our best models are significantly better than a vanilla BERT model, with 89.60\% acc (82.31\% macro $F_1$) for hate speech and 95.20\% acc (70.51\% macro $F_1$) on official TEST data.     
}
\begin{document}
\maketitleabstract

\setcode{utf8}
\setarab

\section{Introduction}
Social media are widely used at a global scale. Communication between users from different backgrounds, ideologies, preferences, political orientations, etc. on these platforms can result in tensions and use of offensive and hateful speech. This negative content can be very harmful, sometimes with real-world consequences. For these reasons, it is desirable to control this type of uncivil language behavior by detecting and removing this destructive content. \\

Although there have been a number of works on detecting offensive and hateful content in English (e.g.~\cite{agrawal2018deep,badjatiya2017deep,nobata2016abusive}), works on many other languages are either lacking or rare. This is the case for Arabic, where there have been only very few works (e.g., ~\cite{alakrot2018towards,albadi2018they,mubarak2017abusive,mubarak2019arabic}). For these motivations, we participated in the Offensive Language and hate-speech Detection shared task organized with the 4\textsuperscript{th} Workshop on Open-Source Arabic Corpora and Processing Tools Arabic (OSACT4). 
\\



Offensive content and hate speech are less frequent online than civil, acceptable communication. For example, only $~19\%$ and $\sim 5\%$ of the released shared task data are offensive and hate speech, respectively. This is the case in spite of the fact that the data seems to have been collected based on trigger seeds that are more likely to accompany this type of harmful content. As such, it is not easy to acquire data for training machine learning systems. For this reason, we direct part of our efforts to automatically augmenting training data released by the shared task organizers (Section ~\ref{subsec:data_aug}). Our experiments show the utility of our data enrichment method. In addition, we hypothesize trained affective models can have useful representations that might be effective for the purpose of detecting offensive and hateful content. To test this hypothesis, we fine-tune one sentiment analysis model and one emotion detection model on our training data. Our experiments support our hypothesis (Section~\ref{sec:model}). All our models are based on the Bidirectional Encoder from Transformers (BERT) model. Our best models are significantly better than competitive baseline based on vanilla BERT. Our contributions can be summarized as follows:

\begin{itemize}
    \item We present an effective method for automatically augmenting training data. Our method is simple and yields sizable additional data when we run it on a large in-house collection. 
    
    \item We demonstrate the utility of fine-tuning off-the-shelf affective models on the two downstream tasks of offensive and hate speech. 
    
    \item We develop highly accurate deep learning models for the two tasks of offensive content and hate speech detection. 
\end{itemize}

The rest of the paper is organized as follows: We introduce related works in Section~\ref{sec:lit}, shared task data and our datasets in Section~\ref{sec:data}, our models in Section~\ref{sec:model}, and we conclude in Section~\ref{sec:conc}.  


\begin{table*}[]
\centering
\begin{tabular}{c|c|c|c|c|c|c|c|c}
\cline{2-9}
                                     &Dataset& \textbf{\#tweets} &\textbf{\# NOT\_OFF} & \textbf{\# OFF}    & \textbf{OFF\%} & \textbf{\# NOT\_HS} & \textbf{\# HS}    &\textbf{ HS \%} \\ \hline
\multirow{3}{*}{}Shard-task data & TRAIN      & 6994     & 5585     & 1409   & 20\%  & 6633    & 361   & 5\%  \\ 
                                     & DEV        & 1000     & 821      & 179    & 18\%  & 956     & 44    & 4\%  \\  
                                     & TEST       & 2000     & -        & -      & -     & -       & -     & -    \\ \hline
\multirow{3}{*}{} Augmented data& AUG-TRAIN-HS & 209780   & -        & -      & -     & 199291  & 10489 & 5\%  \\ 
                                     & AUG-TRAIN-OFF & 480777   & 215365   & 265413 & 55\%  & -       & -     & -    \\ \hline
\end{tabular}
\caption{Offensive (OFF) and Hate Speech (HS) Labels distribution in datasets }
\label{tab:distrib_table}
\end{table*}

\section{Related Work}\label{sec:lit}
\textbf{Thematic Focus:} Research on undesirable content shows that social media users sometimes utilize \textit{profane}, \textit{obscene}, or \textit{offensive} language~\cite{jay2008pragmatics,wiegand2018overview}; \textit{aggression}~\cite{kumar2018benchmarking,modha2018filtering}; \textit{toxic content}~\cite{georgakopoulos2018convolutional,fortuna2018merging,zampieri2019semeval}, and \textit{bullying}~\cite{dadvar2013improving,agrawal2018deep,fortuna2018merging}.\\

\textbf{Overarching Applications:} 
Several works have taken as their target detecting these types of negative content with a goal to build applications for (1) content filtering or (2) quantifying the intensity of polarization~\cite{barbera2015follow,conover2011political}, (3) classifying trolls and propaganda accounts that often use offensive language~\cite{darwish2017seminar}, (4) identifying hate speech that may correlate with hate crimes~\cite{nobata2016abusive}, and (5) detecting signals of conflict, which are often preceded by verbal hostility~\cite{chadefaux2014early}.\\

\textbf{Methods:} A manual way for detecting negative language can involve building a list of offensive words and then filtering text based on these words. As ~\newcite{mubarak2019arabic} also point out, this approach is limited because (1) offensive words are ever evolving with new words continuously emerging, complicating the maintenance of such lists and (2) the offensiveness of certain words is highly context- and genre-dependent and hence a lexicon-based approach will not be very precise. Machine learning approaches, as such, are much more desirable since they are more nuanced to domain and also usually render more accurate, context-sensitive predictions. This is especially the case if there are enough data to train these systems. \\

Most work based on machine learning employs a supervised approach at either (1) character level~\cite{malmasi2017detecting}, (2) word level~\cite{kwok2013locate}, or (3) simply employ some representation incorporating word embeddings~\cite{malmasi2017detecting}. These studies use different learning methods, including Naive Bayes~\cite{kwok2013locate}, SVMs~\cite{malmasi2017detecting}, and classical deep learning such as CNNs and RNNs~\cite{nobata2016abusive,badjatiya2017deep,alakrot2018towards,agrawal2018deep}. Accuracy of the aforementioned systems range between 76\% and 90\%. It is also worth noting that some earlier works~\cite{weber2013secular} use sentiment words as features to augment other contextual features. Our work has affinity to this last category since we also leverage affective models trained on sentiment or emotion tasks. Our approach, however, differs in that we build models free of hand-crafted features. In other words, we let the model learn its representation based on training data. This is a characteristic attribute of deep learning models in general.~\footnote{Of course hand-crafted features can also be added to a representation fed into a deep learning model. However, we do not do this here.} In terms of the specific information encoded in classifiers, researchers use profile information in addition to text-based features. For example, ~\newcite{abozinadah2017detecting} apply SVMs on 31 features extracted from user profiles in addition to social graph centrality measures. \\

Methodologically, our work differs in three ways: (1) we train offensive and hate speech models off affective models (i.e., we fine-tune already trained sentiment and emotion models on both the offensive and hate speech tasks). (2) We apply BERT language models on these two tasks. We also (3) automatically augment offensive and hate speech training data using a simple data enrichment method. \\


\textbf{Arabic Offensive Content:}
Very few works have been applied to the Arabic language, focusing on detecting offensive language. For example,~\cite{mubarak2017abusive} develop a list of obscene words and hashtags using patterns common in offensive and rude communications to label a dataset of 1,100 tweets.~\newcite{mubarak2019arabic} applied character n-gram FasText model on a large dataset 
(3.3M tweets) of offensive content. Our work is similar to~\newcite{mubarak2019arabic} in that we also automatically augment training data based on an initial seed lexicon. \\



\section{Data}\label{sec:data}
In our experiments, we use two types of data: (1) data distributed by the Offensive Language Detection shared task and (2) an automatically collected dataset that we develop (Section~\ref{subsec:data_aug}). The shared task dataset comprises 10,000 tweets manually annotated for two sub-tasks: \textit{offensiveness} (Sub\_task\_A)~\footnote{\url{https://competitions.codalab.org/competitions/22825}.} and \textit{hate speech} (Sub\_task\_B)~\footnote{\url{https://competitions.codalab.org/competitions/22826}}. According to shared task organizers,~\footnote{\url{http://edinburghnlp.inf.ed.ac.uk/workshops/OSACT4/}.},  offensive tweets in the data contain explicit or implicit insults or attacks against other people, or inappropriate language. Organizers also maintain that hate speech tweets contains insults or threats targeting a specific group of people based on the nationality, ethnicity, gender, political or sport affiliation, religious belief, or other common characteristics of such a group. The dataset is split by shared task organizers into 70\% TRAIN, 10\% DEV, and 20\% TEST. Both labeled TRAIN and DEV splits were shared with participating teams, while tweets of TEST data (without labels) was only released briefly before competition deadline.\\

It is noteworthy that the dataset is imbalanced. For \textit{offensiveness} (Sub\_task\_A), only 20\% of the TRAIN split are labeled as offensive and the rest is not offensive. For \textit{hate speech} (Sub\_task\_B), only 5\% of the tweets are annotated as hateful. Due to this imbalanced, the official evaluation metric is macro F\textsubscript{1} score. Table~\ref{tab:distrib_table} shows the size and label distribution in the shared task data splits.~\footnote{Table~\ref{tab:distrib_table} also shows size and class distribution for our automatically extracted dataset, to which we refer to as augmented (AUG).}\\ 

The following are example tweets from the shared task TRAIN split.\\

\textbf{Examples of \textit{offensive} and \textit{hateful} tweets:}

\begin{enumerate}
    \setcounter{enumi}{\value{examples}}
    \item <يا رب يا واحد يا أحد بحق يوم الاحد ان>\\
    < تهلك بني سعود المجرمين. لاجل اطفال اليمن شاركوا.>\\
    \textit{Oh my Lord, O One and Only, destroy the family of Sau`d, for they are the criminals who put children of Yemen to suffer.}~\footnote{Original tweets can be run-on sentences, lack proper grammatical structures or punctuation. In presented translation, for readability, while we maintain the meaning as much as possible, we render grammatical, well-structured sentence.}\\
    \item <يا لبناني يا فضلات الاستعمار الفرنسي>\\
  <اللبنانيي بالخليج يشغلون نسوانهم عاهرات>
  \\
    \textit{Hey, you Lebanese guy, you're the wastes of the French colonizers. The Lebanese in the Gulf put their women in prostitution work.}
\setcounter{examples}{\value{enumi}}\\
\end{enumerate}

\textbf{Examples for \textit{offensive} but \textit{not  hate-speech} tweets:}\\

\begin{enumerate}    
     \setcounter{enumi}{\value{examples}}
     
   \item \<يا لطيف.. يا ساتر ..أحمدوا ربكم إنها >\\
   < مستقعدة كيف لو إنها واقفه>\\
    \textit{Oh my lord... Thank God she has disability. What would have happened if she were not disabled?}\\

    \item \<يا ترى مخبيلنا ايه يا چون سنو انت >
    \\
    <و الكلبوبة اللي جنبك دي>
    \\
    \textit{I wonder what you, and this little pitch by your side, are hiding for us, John Snow?}
    
\setcounter{examples}{\value{enumi}}
\end{enumerate}

\textbf{Examples for \textit{not offensive} and \textit{not  hate-speech} tweets:}\\

\begin{enumerate} 
\setcounter{enumi}{\value{examples}}
    \item 
    <يا بكون بحياتك الأهم يا إما ما بدي أكون>  \\
    \textit{Either I become the most important in your life, or I become nothing at all.}\\
    \item <ايش الاكل الحلو ذا ياسُميه تسلم يدك ياعسل>\\ <ياقشطه ياحلوه ياسُكره يا طباخه يا فنانه يا كل شي>\\
    \textit{Wow! How wonderful this food is, Sumaia! You're such a honey, beauty, sweetie, and good cook! You're are artist! You're everything!}
    \setcounter{examples}{\value{enumi}}
    
\end{enumerate}

\begin{table*}[]
\centering
\begin{tabular}{cl|cl}
\hline
\textbf{Arabic Offensive}     & \textbf{English}                                        & \textbf{Arabic Hateful}          & \textbf{English}                                                  \\ \hline
<يا خييث>   &
You, fat ass!     
& <يا منجاوي>       
& You're Manjawi                \\
<يا رعاع>   &
You're mobby     & 
<يا دندرواي>      
& You're Dandarawi              \\
<يا متشرد>  &
You're a tramp      
& <يا سعوديين>      
& You Saudis               \\ 
<يا مجانين> 
& You're crazy      
& <يا دحباشي>      
& You're Dahbashi              \\ 
<يا جعان>  
& You, hungry man!      
& <يا ادعائي>     
& You, false claimer             \\ 
<يا فاجره>
& You, morally loose     
& <يا حوثي>        
& You, Houthi               \\ 
<يا شمال>  
& Oh, whore     
& <يا شيعي>        
& You, Shiite               \\ 
<يا زباله>
& You, junky       
& <يا عميل>       
& You, spay               \\ 
<يا بهاايم>
& You, animals   
& <يا اخوانجي>      
& You, Ikhwangis             \\
<يا بغيض>   &
You, hateful    
& <يا اخوان>      
& You, Ikhwan             \\ 
<يا وسخه>   
& You, dirty woman      
& <يا ابن الجرابيع> 
& You, son of tramps            \\ 
<يا طاغيه>  
& You, tyrant     
& <يا ابن الحرام>
& You, bastard    \\ 
<يا فاجر>  
& You, salacious 
& <يا ابن الزنا> 
& You, bastard      \\ 
<يا مغفل>  
& You, idiot    
& <يا ابن اليهوديه>
& You, son of Jewish woman       \\ 
<يا رخمه>   &
You, silly woman   
& <يا ابن الزانيه>
& You, son of adulterous woman\\ 
<يا نحس>    &
You, sinister      
& <يا ابن الموهومه>
& You, son of deceived woman     \\ 
<يا غباء>   
& You, stupid head    &
<يا ابن العرص>    
& You, son of pimp  \\ 
<يا كئيب>   
& You, gloomy head   
& <يا ابن المتناكه>
& You, son of adulterous         \\
<يا مرا>    
& You, unworthy woman     &
<يا امارات>       
& You, Emirate             \\ 
<يا حمقي>   
& You, fools    
& <يا اتحادي>     
& You, Itihadi              \\ \hline
\end{tabular}
\caption{Examples of offensive and hateful seeds in our lexica}
\label{tab:OFF_HS_seed}
\end{table*}

\begin{table}[]
\centering
\begin{tabular}{c|l}
\hline
\textbf{Arabic Non-OFF/Non-HS} & \textbf{English} \\\hline
<يا ابطال>   
& You, heros    \\
<يا فنان>    
& You, artist    \\
<يا عبسميع>  
& You, Absemee'    \\
<يا عاالم>   
& Oh, people     \\
<يا منسي>    
& Oh, forgotten man\\
<يا ماعندك>  
& You have a lot     \\
<يا ماما>    
& Oh, mum                                                    \\ 
<يا قمرر>    
& You, beautiful lady      \\
<يا جامد>    
& You, wonderful man     \\
<يا طفل>      
& Oh, child     \\
<يا قطه>     
& Oh, delicate lady       \\
<يا حيلتها> 
& You, lulled      \\
<يا ايتها> 
& Oh you\dots     \\ 
<يا راعي>    
& You, caregiver  \\
<يا حبيبي>   
& Oh, darling  \\ 
<يا رب>       
& Oh, my Lord      \\
<يا واحد>    
& Oh, the One       \\
<يا ناس>    
& You, people    \\
<يا اخر>     
& Oh, the Last     \\
<يا بابا>   
& Oh, daddy                                                   \\
\hline
\end{tabular}
\caption{Examples of non-offensive/non-hateful seeds filtered out from our lexica.}
\label{tab:Not_OFF_HS_seed}
\end{table}
\subsection{Data Augmentation}~\label{subsec:data_aug}
\begin{figure}[h]
\includegraphics[width=8cm]{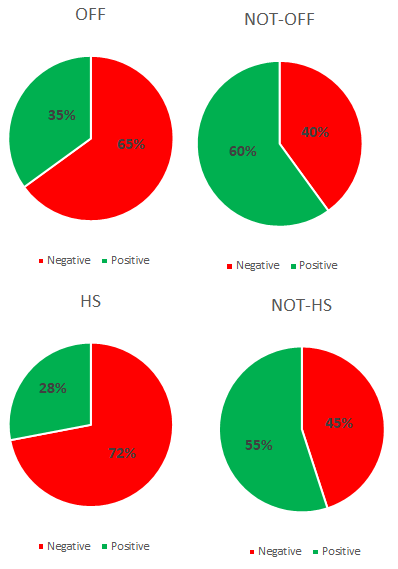}
\caption{Distribution of Negative and Positive Tweets after applied AraNet on Shared-Task TRAIN Data }
\label{img:aranet}
\end{figure}

As explained earlier, the positive class in the offensive sub-task (i.e., the category `offensive') is only 20\% and in the hateful sub-task (i.e., the class `hateful') it is only 5\%. Since our goal is to develop exclusively deep learning models, we needed to extend our training data such that we increase the positive samples. For this reason, we develop a simple method to automatically augment our training data. Our method first depends on extracting tweets that contain any of a seed lexicon (explained below) and satisfy a predicted sentiment label condition. We hypothesize that both offensive and hateful content would carry negative sentiment and so it would be intuitive to restrict any automatically extracted tweets to those that carry these negative sentiment labels. To further test this hypothesis, we analyzing the distribution of the sentiment classes in the TRAIN split using an off-the-shelf tool, \textit{AraNet}~\cite{mageed_osact4}. As shown in Figure \ref{img:aranet}, AraNet assigns sensible sentiment labels to the data. For the `offensive' class, the tool assigns 65\% negative sentiment tags and for the non-offensive class it assigns only 60\% positive sentiment labels.~\footnote{AraNet~\cite{mageed_osact4} assigns only positive and negative sentiment labels. In other words, it does not assign neutral labels.} For the hate speech data, we find that AraNet assigns 72\% negative labels to the `hateful' class and 55\% positive sentiment labels for the `non-hateful' class. Based on this analysis, we decide to impose a sentiment-label condition on the automatically extended data as explained earlier. In other words, we only choose `offensive' and `hateful' class data from tweets predicted as negative sentiment. Similarly, we only choose `non-offensive' and `non-hateful' tweets assigned positive sentiment labels by AraNet. We now explain how we extend the dataset. We now explain our approach to extract tweets with an offensive and hateful seed lexicon. \\

To generate a seed lexicon, we extract all words that follow the \textit{Ya} (\textit{Oh, you}) in the shared task TRAIN split positive class in the two sub-tasks (i.e., `offensive' and `hateful'). The intuition here is that the word \textit{Ya} acts as a trigger word that is likely to be followed by negative lexica. This gives us a set of 2,158. We find that this set can have words that are neither offensive nor hateful outside context and so we manually select a smaller set of 352 words that we believe are much more likely to be effective offensive seeds and only 38 words that we judge as more suitable carriers of hateful content. Table~\ref{tab:OFF_HS_seed} shows samples of the offensive and hateful seeds. Table~\ref{tab:Not_OFF_HS_seed} shows examples of seeds in our initial larger set that we filtered out since these are less likely to carry negative meaning (whether offensive or hateful).\\

To extend the offensive and hateful tweets, we use 500K randomly sampled, unlabeled, tweets from~\cite{abdul2019dianet} that each have at least one occurrence of the trigger word \textit{Ya} and at least one occurrence of a word from either of our two seed lexica (i.e., the offensive and hateful seeds).~\footnote{The 500K collection is extracted via searching a larger sample of $\sim 21M$ tweets that all have the trigger word \textit{Ya}. This corpus is also taken from~\cite{abdul2019dianet}. Note that a tweet can have both an offensive and a hateful seed.} We then apply AraNet \cite{mageed_osact4} on this 500K collection and keep only tweets assigned negative sentiment labels. Tweets that carry offensive seeds are labeled as `offensive' and those carrying hateful seeds are tagged as `hateful'.
This gives us 265,413 offensive tweets and 10,489 hateful tweets. For reference, the majority (\%=67) of the collection extracted with our seed lexicon are assigned negative sentiment labels by AraNet. This reflects the effectiveness of our lexicon as it matches our observations about the distribution of sentiment labels in the shared task TRAIN split.\\

To add positive class data (i.e., `not-offensive' and `not-hateful') to this augmented collection, we randomly sample another 500K tweets that carry \textit{Ya} from ~\cite{abdul2019dianet} that do not carry any of the two offensive and hateful seed lexica. We apply AraNet on these tweets and keep only tweets assigned a positive sentiment label (\%=70). We use 215,365 tweets as `non-offensive' but only 199,291 as `non-hateful'.~\footnote{We decided to keep only 199,291 `non-hateful' tweets since our augmented `hateful' class comprises only 10,489 tweets.} Table~\ref{tab:distrib_table} shows the size and distribution of class labels in our extended dataset.\\

\begin{figure}[h]
\includegraphics[width=8cm]{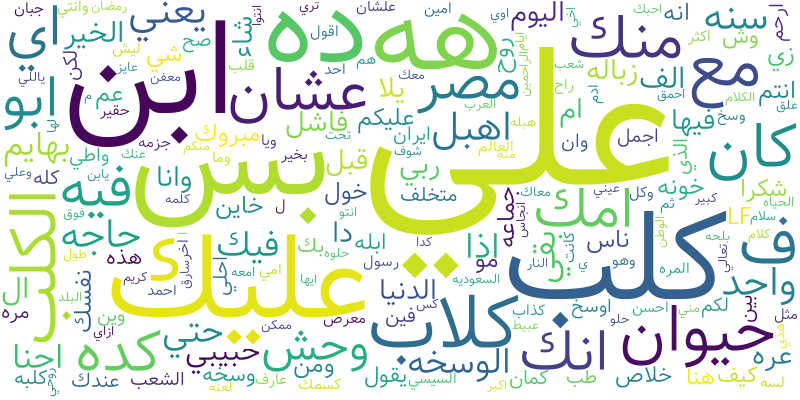}
\caption{A word cloud of unigrams in our extended training offensive data (AUG-TRAIN-OFF).}
\label{img:AUG_OFF_ONLY}
\end{figure}

\begin{figure}[h]
\includegraphics[width=8cm]{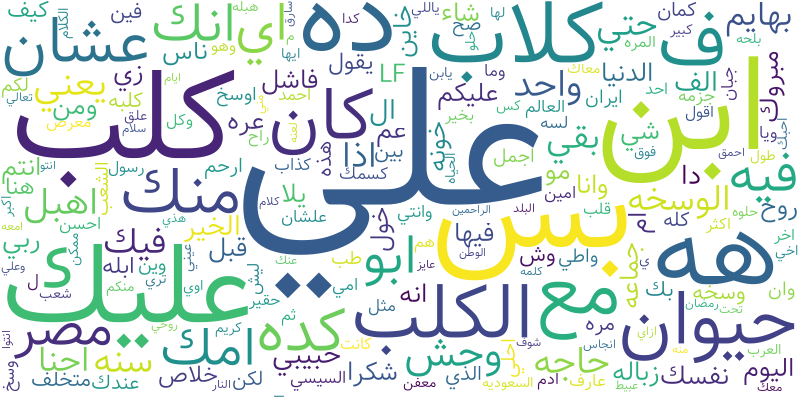}
\caption{A word cloud of unigrams in our extended training hate speech data (AUG-TRAIN-HS).}
\label{img:AUG_HS_ONLY}
\end{figure}

Figure~\ref{img:AUG_HS_ONLY} and Figure~\ref{img:AUG_OFF_ONLY} are word clouds of unigrams in our extended training data (offensive and hateful speech, respectively) after we remove our seed lexica from the data. The clouds show that the data carries lexical cues likely to occur in each of the two classes (offensive and hateful). Examples of frequent words in the offensive class include \textit{dog, animal, son of, mother, dirty woman, monster, mad}, and \textit{on you}. Examples in the hateful data include \textit{shut up, dogs, son of, animal, dog, haha}, and \textit{for this reason}. We note that the hateful words do not include direct names of groups since these were primarily our seeds that we removed before we prepare the word cloud. Overall, the clouds provide sensible cues of our phenomena of interest across the two tasks.

\section{Models}~\label{sec:model}

\begin{table*}[]
\centering
\begin{tabular}{lcc|cc|cc|cc}
\cline{2-9}
\multirow{2}{*}{}          & \multicolumn{4}{c|}{\textbf{Dev}}                                    & \multicolumn{4}{c}{\textbf{Test}}                                \\ \cline{2-9} 
                           & \multicolumn{2}{c}{\textbf{OFF}} & \multicolumn{2}{c|}{\textbf{HS}} & \textbf{OFF}   & \textbf{}      & \textbf{HS}    & \textbf{}      \\ \hline
\textbf{Model}             & \textbf{Acc}    & \textbf{F1}     & \textbf{Acc}    & \textbf{F1}    & \textbf{Acc}   & \textbf{F1}    & \textbf{Acc}   & \textbf{F1}    \\ \hline
\textbf{BERT}        & 87.10           & 78.38           & \textbf{95.70}  & \textbf{70.96} & 87.30          & 77.70          & \textbf{95.20}           & \textbf{70.51}           \\ \hline
\textbf{BERT-SENTI} & 87.40           & 78.84           & 95.50           & 68.01          & 87.45          & 80.51          & 93.15          & 61.57          \\ 
\textbf{BERT-EMO}   & 88.30           & 80.39           & 95.40           & 68.54          & --          & --          & -- & -- \\ \hline
\textbf{BERT-EMO-AUG}   & \textbf{89.60}  & \textbf{82.31}  & 93.90           & 62.52          & \textbf{89.35} & \textbf{82.85} & --             & --             \\ \hline
\end{tabular}
\caption{Offensive (OFF) and Hate Speech (HS) results on DEV and TEST datasets }
\label{tab:results_table}
\end{table*}

\subsection{Data Pre-Processing}

We perform light Twitter-specific data cleaning (e.g., replacing numbers, usernames, hashtags, and hyperlinks by unique tokens NUM, USER, HASH, and URL respectively). We also perform Arabic-specific normalization (e.g., removing diacritics and mapping various forms of Alef and Yeh each to a canonical form). For text tokenization, we use byte-pair encoding (PBE) as implemented in Multilingual Cased BERT model.\\

\subsection{BERT}

Our experiments are based on BERT-Base Multilingual Cased model released by~\cite{devlin2018bert}~\footnote{\url{https://github.com/google-research/bert/blob/master/multilingual.md}.}. BERT stands for \textbf{B}idirectional \textbf{E}ncoder \textbf{R}epresentations from \textbf{T}ransformers. It is an approach for language modeling that involves two self-supervised learning tasks, (1) masked language models (MLM) and (2) next sentence predication (NSP). BERT is equipped with an Encoder architecture which naturally conditions on bi-directional context. It randomly masks a given percentage of input tokens and attempts to predict these masked tokens. ~\cite{devlin2018bert} mask 15\% of the tokens (the authors use \textit{word pieces}) and use the hidden states of these masked tokens from last layer for prediction. To understand the relationship between two sentences, the BERT also pre-trains with a binarized NSP task, which is also a type of self-supervises learning. For the sentence pairs (e.g., \textit{A}-\textit{B}) in pre-training examples, 50\% of the time \textit{B} is the actual next sentence that follows A in the corpus (positive class) and 50\% of the time \textit{B} is a random sentence from corpus (negative class). Google's pre-trained BERT-Base Multilingual Cased model is trained on 104 languages (including Arabic) with 12 layers, 768 hidden units each, 12 attention heads. The model has 119,547 shared word pieces vocabulary, and was pre-trained on the entire Wikipedia for each language. \\



In our experiments, we train our classification models on BERT-Base Multilingual Cased model. 
For all of our fine-tuning BERT models, we use a maximum sequence size of 50 tokens and a batch size of 32. We add a `[CLS]' token at the beginning of each input sequence and, then, feed the final hidden state of `[CLS]' to a Softmax linear layer to get predication probabilities across classes. We set the learning rate to $2e-6$ and train for 20 epochs. We save the checkpoint at the end of each epoch, report F1-score and accuracy of the best model, and use the best checkpoint to predict the labels of the TEST set. We fine-tune the BERT model under five settings. We describe each of these next.\\

\textbf{Vanilla BERT:} We fine-tune BERT-Base Multilingual Cased model on TRAIN set of offensive task and hate speech task respectively. We refer these two models to \textit{BERT}. The offensive model obtains the best result with 8 epochs. As Table~\ref{tab:results_table} shows, for \textit{offensive language} classification, this model obtains 87.10\% accuracy and 78.38 $F_1$ score on DEV set. We submit the TEST prediction of this model to the shared task and obtain 87.30\% accuracy and 77.70 $F_1$ on the TEST set. The \textit{hate speech} model obtains best result (accuracy = 95.7-\%, $F_1$ = 70.96) with 6 epochs. \\

\textbf{BERT-SENTI}
We use a BERT model fine-tuned with on binary Arabic sentiment dataset as released by~\cite{mageed_osact4}. We use this off-the-shelf (already trained) model to further fine-tune on offensive and hate speech tasks, respectively. We replace the Softmax linear layer for sentiment classification with a randomly initialized Softmax linear layer for each task. We refer to these two models as BERT-SENTI. We train the BERT-SENTI models on the TRAIN sets for offensive and hate speech tasks respectively. On $F_1$ score, BERT-SENTI is 0.3 better than vanilla BERT on the offensive task, but 2.95 lower (than vanilla BERT) on the hate speech task. We submit the TEST predictions of both tasks. The offensive model obtain 87.45\% accuracy and 80.51 $F_1$ on TEST.  The hate speech model acquire 93.15\% accuracy and 61.57 $F_1$ on TEST. \\ 

\textbf{BERT-EMO}
Similar to BERT-SENTI, we use a BERT model trained on 8-class Arabic emotion identification from~\cite{mageed_osact4} to fine-tune on the offensive and hate speech tasks, respectively. We refer to this setting as BERT-EMO. We train the models on the TRAIN sets for both offensive and hate speech tasks for 20 epochs. The \textit{offensive} model obtains its best result (accuracy = 88.30\%, $F_1$ = 80.39) with 11 epochs. The \textit{hate speech} model acquires its best result (accuracy = 95.40\%, $F_1$ = 68.54) also with 11 epochs. We do not submit an BERT-EMO on the hate speech task TEST set.\\



\textbf{BERT-EMO-AUG}
Similar to BERT-EMO, we also fine-tune the emotion BERT model (BERT-EMO) with the augmented offensive dataset (AUG-TRAIN-OFF) and augmented hate speech dataset (AUG-TRAIN-HS). On the DEV set, the \textit{offensive model} acquires its best result (accuracy = 89.60\%, $F_1$ = 82.31) with 13 epochs. The best results for the \textit{hate speech model} (accuracy = 93.90\%, $F_1$ = 62.52) is obtained with 9 epochs. Our best offensive predication on TEST is BERT-EMO-AUG. It which achieves an accuracy of 89.35\% and $F_1$ of 82.85. We do not submit an BERT-EMO-AUG on the hate speech task TEST set.

\section{Conclusion}\label{sec:conc}
We described our submission to the offensive language detection in Arabic shared task. We offered a simple method to extend training data and demonstrated the utility of such augmented data empirically. We also deploy affective language models on the two sub-tasks of offensive language detection and hate speech identification. We show that fine-tuning such affective models is useful, especially in the case of offensive language detection. In the future, we will investigate other methods for improving our automatic offensive and hateful language acquisition methods. We also explore other machine learning methods on the tasks. For example, we plan to investigate the utility of semi-supervised methods as a vehicle of improving our models.

\section{Bibliographic References}

\bibliography{lrec2020}
\bibliographystyle{lrec}

\end{document}